\documentclass{article}
\PassOptionsToPackage{numbers, compress}{natbib}

\usepackage[preprint]{neurips_data_2024}

\usepackage[utf8]{inputenc} 
\usepackage[T1]{fontenc}    
\usepackage{hyperref}       
\usepackage{url}            
\usepackage{booktabs}       
\usepackage{amsfonts}       
\usepackage{siunitx}
\usepackage{nicefrac}       
\usepackage{microtype}      
\usepackage{xcolor}         
\usepackage{microtype}
\usepackage{graphicx}
\usepackage{amsmath}
\usepackage{amsthm}
\usepackage{subfigure}
\usepackage{booktabs} 
\usepackage{booktabs}
\usepackage{multirow}
\usepackage{color}
\usepackage{algorithm}
\usepackage{algorithmic}
\usepackage[algo2e]{algorithm2e}
\usepackage{wrapfig}

\newcommand{\model}{\sc ATOM}
\title{From Tissue Plane to Organ World: A Benchmark Dataset for Multimodal Biomedical Image Registration using Deep Co-Attention Networks}

\author{%
  Yifeng Wang\thanks{This work was done when Yifeng was a visiting scholar at UIUC.} \\
  Department of \\Electrical and Computer Engineering\\
  Carnegie Mellon University\\
  Pittsburgh, PA 15213 \\
  \texttt{yifengw3@andrew.cmu.edu} \\
  \And
  Weipeng Li \\
  Department of \\Food Science and Nutrition  \\
  The Hong Kong Polytechnic University 
  \\Hong Kong \\
  \texttt{weipeng.li@polyu.edu.hk} \\
   \AND
   Thomas Pearce \\
   Department of Pathology \\
   University of Pittsburgh \\
   Pittsburgh, PA 15213 \\
   \texttt{pearcetm@upmc.edu} \\
   \And
   Haohan Wang\thanks{Corresponding author} \\
   School of Information Sciences \\
   University of Illinois at Urbana-Champaign \\
   Illinois, IL 61820\\
   \texttt{haohanw@illinois.edu} \\
}

\begin{document}

\maketitle

\begin{abstract}
\label{sec:abstract}
Correlating neuropathology with neuroimaging findings provides a multiscale view of pathologic changes in the human organ spanning the meso- to micro-scales, and is an emerging methodology expected to shed light on numerous disease states.
To gain the most information from this multimodal, multiscale approach, it is desirable to identify precisely where a histologic tissue section was taken from within the organ in order to correlate with the tissue features in exactly the same organ region.
Histology-to-organ registration poses an extra challenge, as any given histologic section can capture only a small portion of a human organ.
Making use of the capabilities of state-of-the-art deep learning models, we unlock the potential to address and solve such intricate challenges. 
Therefore, we create the {\model} benchmark dataset, sourced from diverse institutions, with the primary objective of transforming this challenge into a machine learning problem and delivering outstanding outcomes that enlighten the biomedical community.
The performance of our RegisMCAN model demonstrates the potential of deep learning to accurately predict where a subregion extracted from an organ image was obtained from within the overall 3D volume.
The code and dataset can be found at:
\href{https://github.com/haizailache999/Image-Registration/tree/main}{https://github.com/haizailache999/Image-Registration/tree/main} 

\end{abstract}    
\section{Introduction}
\label{sec:intro}
In medical image processing and analysis, 2D medical tissues are crucial in image-guided surgery, computer-aided detection, and medical data visualization~\cite{litjens2017survey,canalini2019segmentation,ruikar2019automated}.
These tissues allow physicians to visually analyze and examine anatomical structures, pathological regions, or organs with greater clarity.
In recent research, an emerging method that can correlate neuropathology with neuroimaging findings has drawn a lot of attention~\cite{faigle2023brainbox,matsuda2017ex}.
The multiscale view of pathologic changes will disclose the disease states and guide the next step in cure.
Radiology (MRI, CT) provides a view of the tissue which can be used to predict disease. Pathology slides provide a much higher-resolution view of a smaller portion of the tissue and provide different information. Both are useful for making a diagnosis and they are complementary. Registering the pathology image to the radiology image is useful for research purposes because it will provide precise information about the pathologic state of the tissue that corresponds with a radiologic signal, to better understand the basis of the radiologic signal. This is expected to lead to an improved ability to predict disease based on non-invasive radiology studies.

However, this will not be that easy because it needs the accurate position where the histologic tissue section was taken from within the organ.
To gain comprehensive information from the multimodal method, such a position will be crucial in order to correlate the tissue and organ features together.
Multimodal image registration is often performed as a two-step process, where the moving image is first transformed into the same feature space as the target image, and then the features are optimally aligned.
The feature of any given 2D tissue plane can only capture a small portion of the original 3D organ volume, therefore, after transforming from histologic to MRI feature space, the resulting 2D image must be localized within the much larger 3D MRI volume.
Moreover, these 2D tissue planes are usually extracted randomly from \textbf{any} position and \textbf{any} view direction.
These challenges will make it impossible for physicians to directly align the tissue back to the organ precisely. 
At the same time, traditional methods cannot solve this problem as well, considering the complex data format of 3D organ volumes and various types of 2D tissue slices.

Given wide variations in pathology and the potential fatigue of human experts, researchers and physicians have begun to benefit from computer-assisted interventions, especially deep learning~\cite{litjens2017survey,shen2017deep}. 
Deep learning methods can learn the features from both 2D tissue planes and 3D volumetric cellular structures, and then discover their intrinsic relationships, enabling the resolution of this problem a reality.
However, transforming the inherent challenges of this real-world problem into a deep learning problem poses difficulties, particularly in light of the fact that contemporary deep learning models typically require large datasets for effective training.

\begin{figure}
\centering
\includegraphics[width=0.9\linewidth]{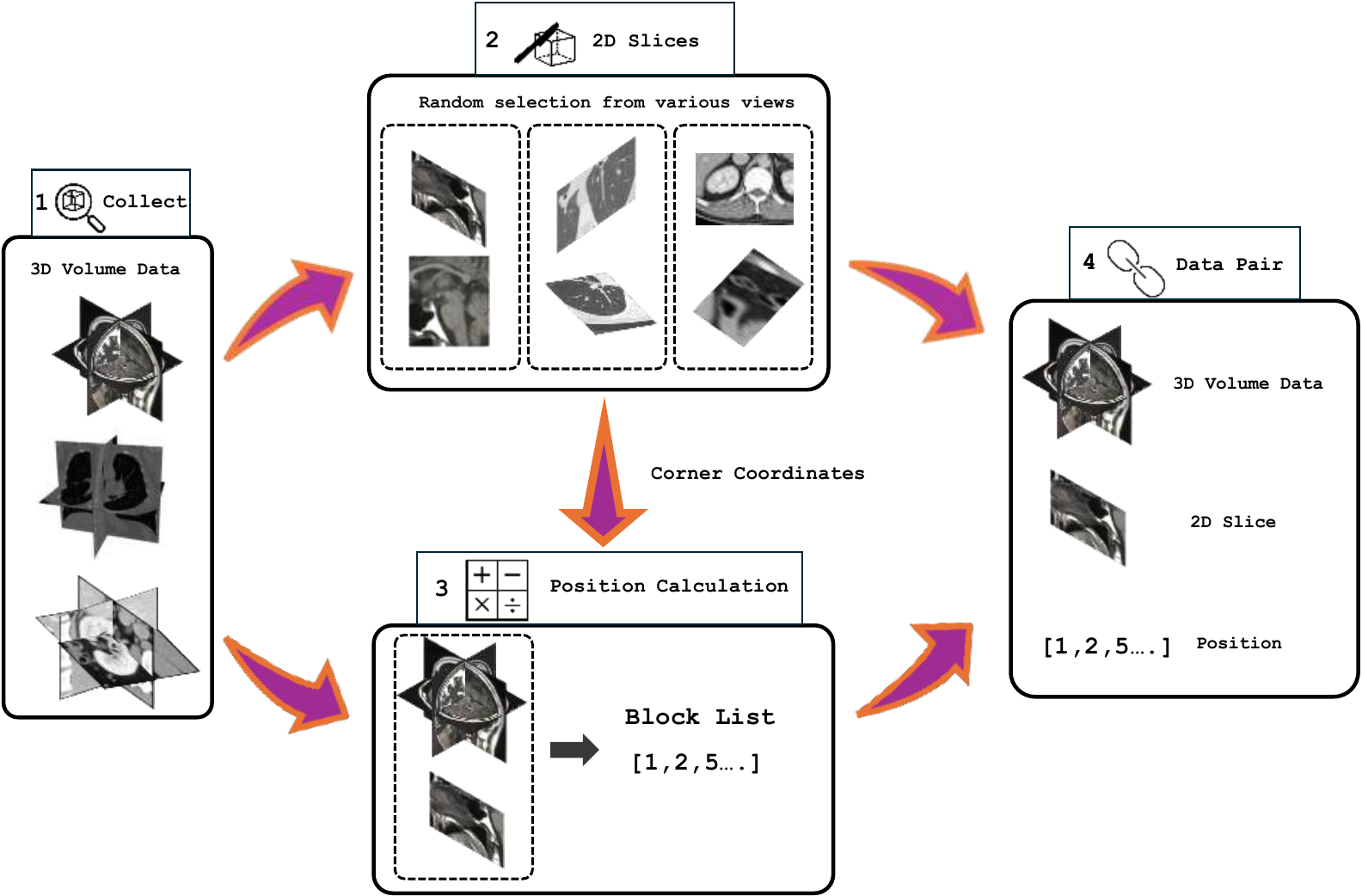}
  \caption{\textbf{Overview of {\model} curation pipeline.} We collect relevant 3D histopathology volume data in \textbf{Collect}. For \textbf{2D Slices} extraction, we randomly cut the 2D slices from different views (not only from axial, coronal, and sagittal views) within the collected 3D volume data and record the corresponding coordinates for each corner. In \textbf{Position Calculation} section, we first cut the 3D volume data into multiple small blocks, and then rely on a conventional calculation model to judge which block the 2D slice belongs to. Finally, relevant 3D volume Data, 2D cell slices and positions are paired up to form \textbf{Data Pairs}, which can yield {\model} dataset, a richly annotated image registration dataset for histopathology.}
  \label{fig:teaser}
\end{figure}

In this paper, we present {\model}, a first attempt to \textbf{a}lign \textbf{t}issue plane back to \textbf{o}rgan volu\textbf{m}e. 
{\model} extends the real-world registration problem to the deep learning domain for end-to-end training of an intelligent biomedical registration agent. It aims to train an algorithm to predict where a subregion extracted from an MRI image was obtained from within the overall 3D volume.
{\model} consists of two main components. 
First, {\model} collects 3D medical volume data from various institutions, including brain MRI images, chest CT, etc. 
To ensure the collected data matches the reality accurately, we employ a random selection process to gather 2D slices from various angles.
Second, inspired by the visual question answering method~\cite{antol2015vqa}, we treat the 3D organ volumes as the type of visual image, 2D tissue slices as a question query, and finally the localized position as the final answer.
{\model} adapts the deep modular co-attention networks~\cite{yu2019deep} to tackle this challenge and bring insights to both biomedical and machine learning communities.

Specifically, our paper makes the following contributions:
\begin{itemize}
    \item \textit{Computational Problem Setting}. We first bring the real-world tissue-organ registration problem into a computational-based problem and bring a linkage from intelligent agents toward solving the biomedical difficulty.
    \item \textit{{\model} Dataset}. We present a novel data generation pipeline to create diverse (2D image, 3D image, output position) instances, by sampling 2D cell slices from collected 3D volume biomedical data to create intended output positions. This requires zero manual annotations and creates the \textbf{first} diverse biomedical image registration dataset similar to reality by packing up the multi-modal (2D, 3D) image-position pairs.

    \item \textit{RegisMCAN}. We propose a novel learning method for successfully adapting MCAN~\cite{yu2019deep} to the biomedical image registration domain using our self-generated biomedical multi-modal image-position dataset. Specifically, we treat the 3D biomedical volume data as the image format in traditional VQA problem and extract relative features. We then continue to transform the 2D cell slices into question format and extract the correlative information. Our empirical study validates the effectiveness of the deep modular co-attention method and reveals best practices and interesting findings for adapting the real-world biomedical image pairs in such a way. 

\end{itemize}
\section{Background}
\label{sec:Related}



The preparation of 2D tissue section often undergoes sampling of tissue, fixation, dehydration, embedding, sectioning, mounting, and staining~\cite{liu2021harnessing}. 
The preparation of tissues often involves destructive methods.
For example, in brain section preparation, due to the large scale of the brain, it may be processed to small volumes of interest.
The embedded tissue is sliced using a microtome, producing thin sections ranging from 5 to 20 micrometers, where the fragmented 2D tissue slices are difficult to realign to 3D volume~\cite{faigle2023brainbox}.
Through histological analysis, the professionals can diagnose the targeted diseases.
For instance, nonalcoholic steatohepatitis (NASH) can be diagnosed and classified as mild, moderate, and severe based on the degree of steatosis, ballooning and disarray, and inflammation~\cite{brunt1999nonalcoholic,takahashi2014histopathology}.

\begin{wrapfigure}{r}{0.5\textwidth}
\includegraphics[width=0.9\linewidth]{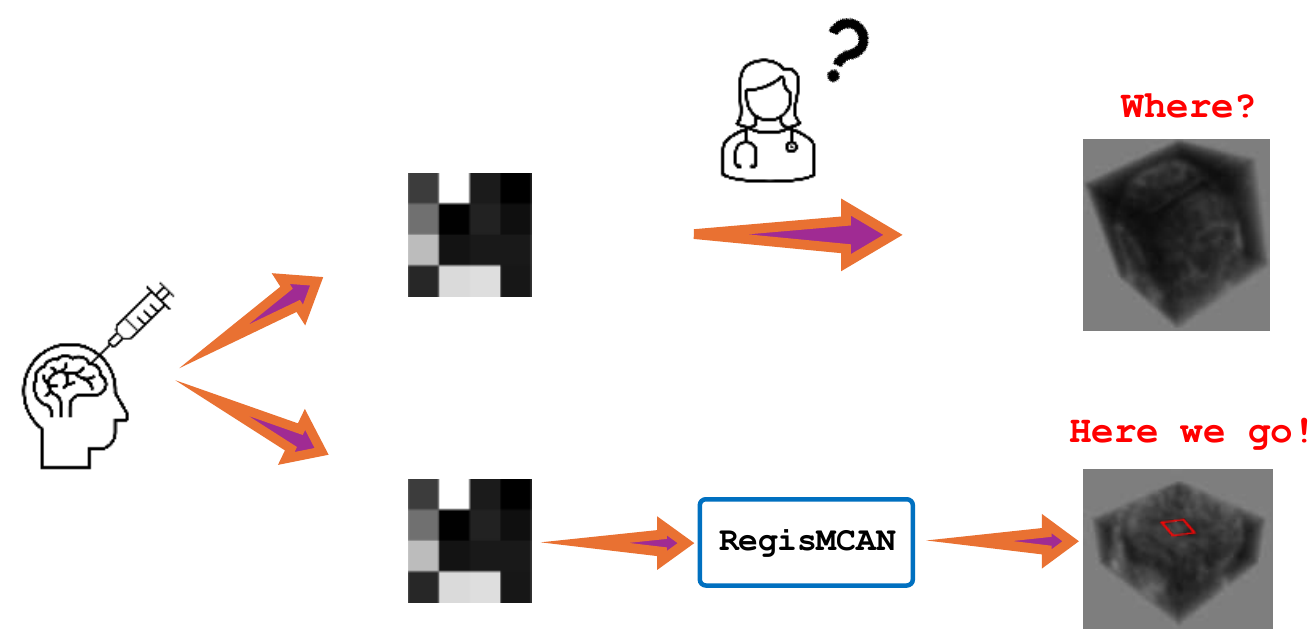} 
\caption{Visualization of the tissue-MRI registration problem.
}
\label{fig:bg}
\end{wrapfigure}

Throughout the processes mentioned above, the tissue that has been sectioned undergoes alterations in comparison to its original stage in living organisms (\textit{in vivo} state). These changes encompass shrinkage and distortion caused by fixation and embedding, tearing of tissue sections, misalignment of tissue sections for microscopic observation, misarrangement of tissue sections to fit standard microscope slides, mismatch in slice thickness between sections prepared for light microscopy, as well as compression and uneven distortion during the sectioning procedure~\cite{alyami2022histological}.
Considering that the 2D tissue usually can only capture a small portion of the original 3D volume, these changes will finally cause a feature unalignment problem and cause difficulty in accurately linking histology back to 3D organ imaging. Fig~\ref{fig:bg} shows the visualization of the problem and the 2D tissue usually cannot be successfully aligned easily even by physicians because it only has a small portion of original 3D organ features.
Matching MRI data with histological images enables researchers and clinicians to directly compare \textit{in vivo} imaging with microscopic tissue characteristics and enhance their understanding of the structure, composition, and pathological changes.

Organ imaging techniques have been devised to evaluate tissue composition and diagnose specific pathological conditions. 
However, traditional organ imaging techniques, such as MRI, have limitations for further tissue pathology understanding.
Particularly, multiple sclerosis, including the breakdown of the blood-brain barrier, the infiltration of immune cells, and the destruction of myelin sheaths, is difficult to assess through MRI but can be identified with histology~\cite{filippi2019association,filippi2016mri,lassmann2018multiple}.
That's why linking pathology back to organ image is urgently needed.

Therefore, to have a better understanding of the connection between imaging and histology, there has been an increasing trend of acquiring imaging and histology on the same tissue.
A group of scientists has tried to establish the connection by performing MRI followed by histological analysis of the region of interest in a Brainbox~\cite{faigle2023brainbox}.
However, this method needs to extract the real organ from the human and process the 3D volume and 2D MRI scan at the same time, and the chemical used, formblin, costs more than 500 dollars per liter and may interfere with histological analysis~\cite{faigle2023brainbox}.
At the same time, complex and manual tissue-MRI processing steps will be time and resource-consuming.
Another approach is to employ the free software Fiji (ImageJ) to match histological results and MRI images~\cite{schindelin2012fiji,khodanovich2022challenges}.
Briefly, the pre-processed MRI image with the defined regions of interest (ROIs) is scaled to a higher resolution, followed by the generation of a red-green-blue (RGB) composite image, resulting in highlighted ROIs on the histology image. 
The final step is to scale the histology image to a lower resolution and the final ROIs are obtained by decomposing the composite image into channels~\cite{khodanovich2022challenges}.
The limitations of this method are that it cannot process the images with local imperfections and the correctness largely depends on the experience of the operator~\cite{khodanovich2022challenges}.
So we aim to create an intelligent, fully automated agent that can solve this problem in an easy way. We propose RegisMCAN as the agent and collected {\model} dataset to train this agent. Our {\model} curation pipeline can be easily used by other researchers repeatedly to create larger datasets for future research.

\section{ATOM Dataset Preparation}

However, training an intelligent agent is not that easy.
First, it's impossible for a computational agent to deal with histology tissue and organs of humans directly.
Second, there is a lack of biomedical image registration datasets to train the intelligent agent to help physicians tackle this real-world challenge. 
To fill this gap, we first convert the problem from tissue-organ mapping into the slice-organ mapping problem, note that the slice is directly obtained from the imaging of the organs. 
Then we create the first dataset of its kind from widely existing different kinds of 3D biomedical images, through a fully automated procedure. 
It consists of four main parts, 3D volume data collection, 2D slice gathering, label generation (block calculation) and the data are paired up at the final stage to form the complete datasets.
The 3D volume data collection will collect the 3D volumes simulating the organs of patients.
The 2D slice gathering will collect the 2D slices simulating the tissue from these organs.
The label generation will help to mark down the ground truth and the final paired-up data can be used to train the intelligent agent.

\begin{table*}[ht]\centering
\scalebox{0.9}{
\begin{tabular}{@{}llllll@{}}
\toprule
\textbf{Data Type}        & \textbf{Dataset Name} & \textbf{Source} & \textbf{Training Size}    & \textbf{Validation Size} & \textbf{Testing Size}    \\ \midrule
MRI       & ADNI            & ADNI    & 800 3D                     &  100 3D                        &   100 3D                       \\ \midrule
Abnormal CT               & OrganMNIST3D          & Liver Tumor     & 971 3D  & 161 3D  & 610 3D\\ \midrule
\multirow{2}{*}{Chest CT} & NoduleMNIST3D         & LIDC-IDRI       & 1158 3D  & 165 3D  & 310 3D   \\ \cmidrule(l){2-6} 
          & FractureMNIST3D & RibFrac & 1027 3D& 103 3D  & 240 3D \\ \midrule
Brain MRA & VesselMNIST3D   & IntrA   & 1335 3D & 191 3D & 382 3D\\ \bottomrule
\end{tabular}}
\caption{Data Specifications. Here we only show the size of 3D organ data. For the ADNI dataset, we randomly select 800 samples for training set, 100 for validation and testing sets.}
\label{tab_data}
\end{table*}

\textbf{3D Volume Data Collection.}
We first try to collect different kinds of 3D biomedical volume data from public datasets. 
These datasets contain various formats, including: magnetic resonance imaging (MRI), abdominal computed tomography (CT), chest CT, brain MR angiography (MRA), and Electron Microscope.
These data formats are very popular in nowadays biomedical research fields.
\begin{itemize}
    \item \textit{Magnetic Resonance Imaging.} MRI is usually used in Alzheimer’s Disease (AD) diagnosis from brain imaging. 
    We collect the dataset from the Alzheimer’s Disease Neuroimaging Initiative (ADNI). 
    There are a total of 3,891 3D MRI images in the dataset, including 1,216 normal cases (NC), 1,110 AD cases and 1,565 Mild Cognitive Impairment (MCI) cases. MCI is a kind of middle stage.

    \item  \textit{Abdominal CT.} Abdominal CT has been widely used as an imaging tool to assess liver morphology, texture, and focal lesions, which is important in liver cancer research.
    We collect the organ related CT directly from MedMNIST~\cite{medmnistv2}, whose source data is from Liver Tumor Segmentation Benchmark~\cite{bilic2023liver}. 
    This dataset contains 1,742 samples and is divided into 971 / 161 / 610 for training / validation / test respectively.
    
    \item \textit{Chest CT.} Chest radiography is a widely conducted radiologic examination, which can be used for the identification of lung nodules and diagnosis of rib fractures.
    We collect the NoduleMNIST3D from LIDC-IDRI~\cite{armato2011lung} and FractureMNIST3D from the RibFrac Dataset~\cite{jin2020deep}.
    The 1,633 NoduleNIST3D samples and 1,370 FractureMNIST3D samples are split into 1,158 / 165 / 310 and 1,027 / 103 / 240 according to training / validation / test.

    \item \textit{Brain MR Angiography.} By reconstructing MRA images, 3D models (meshes) of entire brain vessels play a crucial role in intracranial aneurysm diagnosis. 
    Following MedMNIST~\cite{medmnistv2}, the non-watertight mesh is fixed with PyMeshFix~\cite{attene2010lightweight} and the watertight mesh is voxelized with trimesh into 28 × 28 × 28 voxels.
    The original dataset is from IntrA~\cite{yang2020intra}, an open-access 3D intracranial aneurysm dataset.

\end{itemize}

\textbf{2D Slices Gathering.}
For 3D volume data with high resolution, we first downsample the data into low resolution $\textbf{x}_{3D}$ $\in$ $\mathbb{R}^{20\times 20\times 20}$ to prominent pathological features, which is important in deep learning-based medical image analysis~\cite{nirthika2022pooling,mahapatra2019image,wiestler2020deep}.
Then, we extract 2D slices from various perspectives to simulate the diversity of views that physicians encounter, ensuring that the slices are not limited to a single axis.
The 2D slices will have the resolution $\textbf{x}_{2D}$ $\in$ $\mathbb{R}^{4\times 4}$ to keep the particular ratio compared to 3D volumes.
The relative corner coordinates are recorded for block calculation in the next section.

\textbf{Block Calculation.}
In most scenarios, an approximate position will help the physicians correlating the features from histology to organ.  
Considering this, we soften the limitations and allow some of the errors in predicting the final coordinates. 
So we segment the 3D images into small blocks, enabling the model to predict the specific block from which each 2D slice originates.
To further simplify the task, we initially segment the 3D images into several relatively larger blocks, and then further divide each of these larger blocks into multiple smaller ones.
To ensure each 2D slice can be mapped into one block and considering the final accepted error, we use the method of convolution for reference. 
We propose two error resolutions. 
We set the first layer convolution block with high error resolution as the shape of $\textbf{x}_{conv1}$ $\in$ $\mathbb{R}^{10\times 10\times 10}$.
Then we use a step size 5 to move the convolution block, resulting in a total of 27 bigger blocks with the shape equal to $\textbf{x}_{conv1}$.
For each big block we get, we do the same thing with a second layer convolution block in low error resolution $\textbf{x}_{conv2}$ $\in$ $\mathbb{R}^{6\times 6\times 6}$ and step size 2.

This block segmentation step is crucial for making the learning task feasible for a deep learning model. 
In the original configuration, the model is required to predict the coordinates of four corners, each of which corresponds to nearly 8,000 potential points within the 3D volume data's reference frame. Consequently, the likelihood of accurately predicting all corners correctly is less than E-15, making the task impossible for deep learning models.
With the proposed segmentation method, we can successfully improve the possibility to at least 1/27 in high error resolution, and 1/729 in low error resolution settings.

\textbf{Data Pair Up.} After we calculate all the relevant information, we can pair up the data together. A group of data should contain the 3D volume data, the 2D slice, and the block list as labels.
In this case, the total summarization of our {\model} dataset is shown in Table~\ref{tab_data}.

\section{RegisMCAN}
\label{RegisMCAN}

\begin{figure}[h]
\centering
\includegraphics[width=0.98\linewidth]{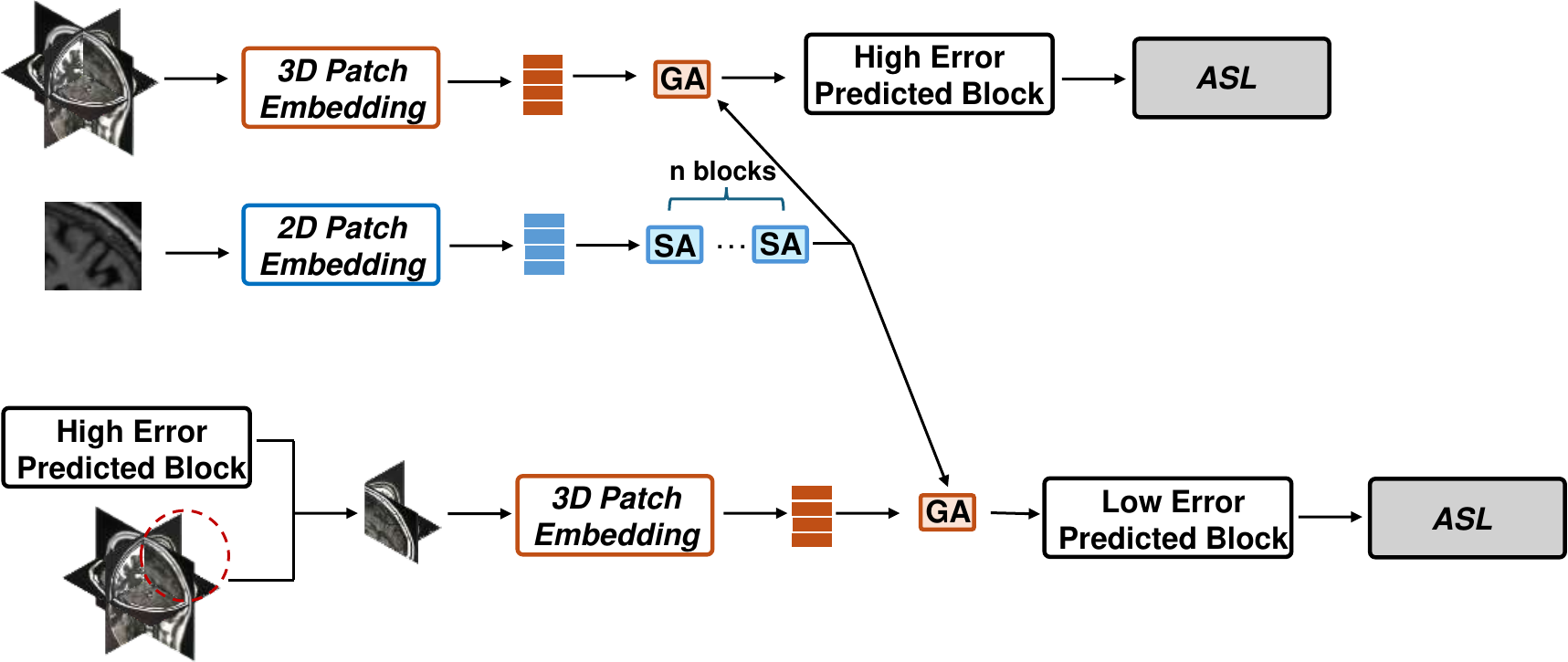}
  \caption{\textbf{Overview of RegisMCAN model.} RegisMCAN consists of two layers. It will first use patch embedding layers to embed the whole 3D volume data and 2D tissue slices separately and use SA and GA units to realize the information extraction and relationship simulation. Here to distinguish the embedding of each separate prediction block, we drop the SA unit and only use one GA unit while realizing relationship simulation.}
  \label{fig:model}
\end{figure}

We employ MCAN~\cite{yu2019deep}, a popular deep learning VQA model, as the foundation model, and continuously adapt the model to our {\model} dataset. 
The traditional MCAN model employs Self-Attention(SA) unit to extract question features and image features separately, and employs Guided-Attention(GA) unit to model the pairwise relationship between each paired QA sample. Then Encoder-Decoder format deep co-attention learning was used to pass forward the features.

We also follow the MCAN, converting 2D slices to Question, and 3D volume data to Answer. 
Here we first reshape the 2D slice $x_{2d}$$\in$$\mathbb{R}^{H\times W\times 1}$ into a sequence of flattened 2D patches $p_{2d}$$\in$$\mathbb{R}^{N\times (P^2\cdot 1)}$, where $(H,W)$ is the resolution of the original image, and the number of channels will be 1 considering the medical slice is grayscale, $(P,P)$ is the resolution of each image patch, and $N = HW/P^2$ is the resulting number of patches.
For 3D volume $x_{3d}$$\in$$\mathbb{R}^{H\times W\times D\times 1}$, where $D$ is the depth of the volume, the flattened 3D patches should be $p_{3d}$$\in$$\mathbb{R}^{N_{3d}\times (P^3\cdot 1)}$. Here to reserve the special features for each high/low error predicted block, we use a 3D convolution based patch embedding method, with $(P,P,P)$ is the resolution of each 3D image patch, and $N = (H-P+S)\cdot(W-P+S)\cdot(D-P+S)/S^3$ is the number of 3D patches. $S$ is the convolutional step size.

We use SA units to extract the features from 2D slices, then pass the related features extracted from SA unit to GA unit.
The SA unit extracts n key-value pairs which are packed into a key matrix $K_{2d}$$\in$$\mathbb{R}^{n\times d}$ and a value matrix $V_{2d}$$\in$$\mathbb{R}^{n\times d}$ from the 2D cell slice. 
Given the query matrix $Q_{3d}$$\in$$\mathbb{R}^{n\times d}$ from the 3D volume data, the feature F$\in$$\mathbb{R}^{n\times d}$ is obtained by performing a weighted summation of all values V, using weights determined by the attention learned from the queries Q and keys K:
\begin{equation}
\label{attention_tradition}
\begin{aligned}
F=softmax(\frac{Q_{3d}K_{2d}^T}{\sqrt{d_k}})V_{2d}
\end{aligned}
\end{equation}
With modeling the relationship in GA unit, we can predict the related high-error big blocks which the 2D slice is aligned with.
Then with the high error predicted block, we can extract another 3D volume block, which will be used to calculate the smaller blocks the 2D slice belongs to. The model feed-forward process is exactly the same and shown in Fig~\ref{fig:model}.

At the same time, considering that the 3D convolution based block segmentation, a 2D cell slice may not only originate from a single block.
We convert the final label into one-hot encoding version and apply Asymmetric Loss~\cite{ridnik2021asymmetric} to help our RegisMCAN converge faster. 
Here, ASL can optimize both correct and incorrect blocks, making it highly suitable for our model given the significantly larger number of negative labels compared to positive ones. The ASL is shown in equation\ref{equ:score}. Here $p\in[0,1]$ is the model’s estimated probability for the class with positive labels originated from Focal Loss~\cite{lin2017focal}. 
$\gamma_{+}$ and $\gamma_{-}$ weight the contribution from positive samples and negative samples. $m>0$ is a probability margin that will fully discard negative samples when their probability is very low.
The probability shifting method can deal with the imbalance in multilabel classification efficiently, making ASL a better fit for our registration challenge.

\begin{equation}
\label{equ:score}
    ASL=
        \begin{cases}
            L_{+}=(1-p)^ {\gamma_{+}} log(p)\\
            L_{-}=(max(p-m,0))^ {\gamma_{-}}log(1-max(p-m,0))
        \end{cases}
\end{equation}

\section{Experiments}

We conduct experiments to study four key problems: 
(1) How do different data types affect the performance of RegisMCAN? (\ref{data_type}) 
(2) How do different data labels (disease or not) affect the performance of RegisMCAN? (\ref{data_label}) 
(3) How do different dataset-splitting methods affect the performance of RegisMCAN? (\ref{data_split})
(4) How do different 2D slice-gathering methods affect the performance of RegisMCAN? (\ref{slice_gather})
All the experiments are conducted on a 40G NVIDIA A100 GPU using Pytorch~\cite{paszke2019pytorch}.
The numerical results are reported with mean and standard deviation in three times.

\subsection{Data Types Effect}
\label{data_type}
\begin{table}[h]
\resizebox{\textwidth}{15.1mm}{
\begin{tabular}{@{}lcccccccccc@{}}
\toprule
\multirow{2}{*}{Data Types} & \multicolumn{1}{l}{\multirow{2}{*}{Slice Gathering}} & \multicolumn{1}{l}{\multirow{2}{*}{Training Size}} & \multicolumn{1}{l}{\multirow{2}{*}{Validation Size}} & \multicolumn{1}{l}{\multirow{2}{*}{Testing Size}} & \multicolumn{3}{c}{High Error Prediction} & \multicolumn{3}{c}{Low Error Prediction} \\ \cmidrule(l){6-11} 
 & \multicolumn{1}{l}{} & \multicolumn{1}{l}{} & \multicolumn{1}{l}{} & \multicolumn{1}{l}{} & Val acc. & Test acc. & Random Guess acc. & Val acc. & Test acc. & Random Guess acc. \\ \midrule
ADNI & Easy & 121600 & 14000 & 15600 & 71.47$\pm$0.32 &70.29$\pm$0.32  & 18.52 &41.95$\pm$0.26 &41.20$\pm$0.41  & 3.43 \\
ADNI & Hard & 121600 & 14000 & 15600 & 71.43$\pm$0.49 &72.76$\pm$0.60  & 18.52 &42.75$\pm$0.43 &44.50$\pm$1.46  & 3.43 \\
OrganMNIST3D & Easy & 147592 & 31072 & 15536 &41.02$\pm$0.05 &38.52$\pm$1.80 & 18.52 &25.5$\pm$1.35 &24.53$\pm$0.96  & 3.43 \\
OrganMNIST3D & Hard & 147592 & 31072 & 15536 &35.47$\pm$0.05 &36.14$\pm$0.38 & 18.52 &20.87$\pm$0.53 &20.39$\pm$0.30 & 3.43 \\
FractureMNIST3D & Easy & 156104 & 32864 & 16432 &41.72$\pm$0.07 &41.29$\pm$0.01 & 18.52 &24.31$\pm$0.10 &23.63$\pm$0.15 & 3.43 \\
FractureMNIST3D & Hard & 156104 & 32864 & 16432 &37.25$\pm$0.08 &37.12$\pm$0.43 & 18.52 &21.36$\pm$0.19 &22.85$\pm$0.41 & 3.43 \\
NoduleMNIST3D & Easy & 176016 & 37056 & 18528 &40.19$\pm$0.05 &40.24$\pm$0.98 & 18.52 &22.99$\pm$0.49 &23.06$\pm$1.05 & 3.43 \\
NoduleMNIST3D & Hard & 176016 & 37056 & 18528 &36.08$\pm$0.34 &35.53$\pm$0.32 & 18.52 &18.87$\pm$0.23 &21.80$\pm$0.17 & 3.43 \\
VesselMNIST3D & Easy & 202920 & 42720 & 21360 &44.64$\pm$0.66 &45.47$\pm$0.84 & 18.52 &24.24$\pm$0.49 &24.17$\pm$0.35 & 3.43 \\
VesselMNIST3D & Hard & 202920 & 42720 & 21360 &41.08$\pm$0.28 &39.06$\pm$0.03 & 18.52 &19.72$\pm$0.35 &22.61$\pm$0.23 & 3.43 \\ \bottomrule
\end{tabular}}
\caption{Comparison of k-5 accuracy ($\%$) with standard deviation among different data types in ATOM. The slice-gathering method has easy and hard ones, details are shown in \ref{slice_gather}.}
\label{exp_q1}
\end{table}

We first test RegisMCAN on various data types in {\model}. We list out the size of training/validation/testing datasets for each data type. As discussed in Section~\ref{RegisMCAN}, we calculate not only the low error prediction accuracy but also the high error prediction accuracy.
This is important because if the model cannot successfully predict the big block, it cannot predict the next step and get the correct results finally.
We also add the column showing the accuracy that without RegisMCAN and the block was randomly guessed.

Results in Table~\ref{exp_q1} show that RegisMCAN achieves the best performance on ADNI. On other organ data, the accuracy is similar. This is reasonable because for the original ADNI data, the images have the shape of $\textbf{x}_{3D}$ $\in$ $\mathbb{R}^{169\times 208\times 179}$. However, for other data, the original images only have the shape of $\textbf{x}_{3D}$ $\in$ $\mathbb{R}^{28\times 28\times 28}$.
Images with higher resolution will have more details and even downsampled to lower resolution, the difference among each block will be kept better.
This kind of inter-block difference will help the model better define the position of the 2D tissue slice and is crucial to get better results.
At the same time, when RegisMCAN can get better performance on the high error prediction, it can also get a better performance on the final result.

\subsection{Data Labels Effect}
\label{data_label}
\begin{table}[h]
\resizebox{\textwidth}{16mm}{
\begin{tabular}{@{}lcccccccl@{}}
\toprule
\multirow{2}{*}{Data Types} & \multicolumn{1}{l}{\multirow{2}{*}{Slice Gathering}} & \multicolumn{1}{l}{\multirow{2}{*}{Training Size}} & \multicolumn{1}{l}{\multirow{2}{*}{Validation Size}} & \multicolumn{1}{l}{\multirow{2}{*}{Testing Size}} & \multicolumn{2}{c}{High Error Prediction} & \multicolumn{2}{c}{Low Error Prediction} \\ \cmidrule(l){6-9} 
 & \multicolumn{1}{l}{} & \multicolumn{1}{l}{} & \multicolumn{1}{l}{} & \multicolumn{1}{l}{} & Val acc. & Test acc. & Val acc. & \multicolumn{1}{c}{Test acc.} \\ \midrule
ADNI-AD & Easy & 121600 & 14000 & 15600 &57.90$\pm$0.44 &53.82$\pm$0.18  &30.14$\pm$0.04 &26.12$\pm$0.34  \\
ADNI-AD & Hard & 121600 & 14000 & 15600 &54.46$\pm$0.06 &51.63$\pm$0.16 &31.36$\pm$0.47 &26.15$\pm$0.66  \\
ADNI-CN & Easy & 121600 & 14000 & 15600 &57.14$\pm$0.46 &53.97$\pm$0.27  &32.25$\pm$0.87 &24.80$\pm$0.50  \\
ADNI-CN & Hard & 121600 & 14000 & 15600 &54.47$\pm$0.28 &51.81$\pm$0.31  &26.08$\pm$0.23 &25.23$\pm$0.43  \\
ADNI-MCI & Easy & 121600 & 14000 & 15600 &57.67$\pm$0.17 &53.93$\pm$0.16  &31.93$\pm$0.13 &28.76$\pm$1.24  \\
ADNI-MCI & Hard & 121600 & 14000 & 15600 &55.24$\pm$0.42 &52.05$\pm$0.38 &25.98$\pm$0.31 &26.61$\pm$0.56  \\
ADNI-ALL & Easy & 121600 & 14000 & 15600 &71.47$\pm$0.32 &70.29$\pm$0.32  &41.95$\pm$0.26 &41.20$\pm$0.41  \\
ADNI-ALL & Hard & 121600 & 14000 & 15600 &71.43$\pm$0.49 &72.76$\pm$0.60  &42.75$\pm$0.43 &44.50$\pm$1.46  \\ \bottomrule
\end{tabular}}
\vspace{0.1cm}
\caption{Comparison of k-5 accuracy (\%) with standard deviation among different data labels. These data are all from ADNI dataset. ADNI-All means the data label has three kinds: AD, CN and MCI. The slice-gathering method has easy and hard ones, details are shown in \ref{slice_gather}.}
\label{exp_q2}
\end{table}

To verify how different data labels will affect the performance of RegisMCAN, we choose out the ADNI dataset from {\model}, and create three variations: ADNI-AD, ADNI-CN, ADNI-MCI. ADNI-AD means all the training, validation and testing samples are from AD images. We randomly select 1000 3D volumes from ADNI and process them with {\model} pipeline. Others variations in a similar way. For ADNI-ALL, it's the same dataset as shown in Table~\ref{exp_q2}.

We find an interesting result that on ADNI-ALL dataset, RegisMCAN has achieved the greatest performance and the result is much higher than other variants. This may due to the comprehensive learning ability of RegisMCAN. 
If the model can be trained with 3D volume data with different labels, it can learn inter- and intra-label features.
This will help to reduce the bias of learning from only similar 3D images. 
Also because CN,MCI,AD are three continuous stages for a patient with Alzheimer's Disease, if the model can be trained with various types of data, it will know better for each small block, what will happen to them next. Focusing on the changes will help to guide the model focusing more on the difference between each small block.

\subsection{Dataset Splitting Effect}
\label{data_split}
\begin{table}
[ht]\resizebox{\textwidth}{35mm}{
\begin{tabular}{lccccccccc}
\hline
\multirow{2}{*}{Data Types} & \multicolumn{1}{l}{\multirow{2}{*}{Slice Gathering}} & \multicolumn{1}{l}{\multirow{2}{*}{Same}} & \multicolumn{1}{l}{\multirow{2}{*}{Training Size}} & \multicolumn{1}{l}{\multirow{2}{*}{Validation Size}} & \multicolumn{1}{l}{\multirow{2}{*}{Testing Size}} & \multicolumn{2}{c}{High Error Prediction} & \multicolumn{2}{c}{Low Error Prediction} \\ \cline{7-10} 
 & \multicolumn{1}{l}{} & \multicolumn{1}{l}{} & \multicolumn{1}{l}{} & \multicolumn{1}{l}{} & \multicolumn{1}{l}{} & Val acc. & Test acc. & Val acc. & Test acc. \\ \hline
OrganMNIST3D & Easy & Pos. & 147592 & 24472 & 92720 &48.41$\pm$0.24 &48.38$\pm$0.36 &26.19$\pm$1.14 &25.55$\pm$0.23  \\
OrganMNIST3D & Easy & Vol. & 147592 & 31072 & 31072 &39.21$\pm$0.55 &49.79$\pm$0.37 &23.66$\pm$0.15 &34.39$\pm$2.73 \\
OrganMNIST3D & Easy & None & 147592 & 24472 & 102480 &41.02$\pm$0.05 &38.52$\pm$1.80 &25.5$\pm$1.35 &24.53$\pm$0.96  \\
OrganMNIST3D & Hard & Pos. & 147592 & 24472 & 92720 &44.13$\pm$0.79 &43.45$\pm$0.28 &22.99$\pm$0.55 &23.87$\pm$0.53  \\
OrganMNIST3D & Hard & Vol. & 147592 & 31072 & 31072 &37.32$\pm$0.12 &48.59$\pm$1.22 &22.53$\pm$0.75 &28.83$\pm$0.48 \\
OrganMNIST3D & Hard & None & 147592 & 24472 & 102480 &35.47$\pm$0.05 &36.14$\pm$0.38 &20.87$\pm$0.53 &20.39$\pm$0.30  \\
FractureMNIST3D & Easy & Pos. & 156104 & 15656 & 36480 &50.96$\pm$0.30 &51.45$\pm$0.38 &29.72$\pm$0.65 &29.45$\pm$0.60  \\
FractureMNIST3D & Easy & Vol. & 156104 & 32864 & 32864 &40.01$\pm$0.73 &46.33$\pm$1.26 &24.79$\pm$0.18 &35.12$\pm$1.57  \\
FractureMNIST3D & Easy & None & 156104 & 15656 & 40320 &41.72$\pm$0.07 &41.29$\pm$0.01 &24.31$\pm$0.10 &23.63$\pm$0.15  \\
FractureMNIST3D & Hard & Pos. & 156104 & 15656 & 36480 &47.24$\pm$0.14 &46.42$\pm$0.25 &27.24$\pm$0.59 &26.52$\pm$0.49  \\
FractureMNIST3D & Hard & Vol. & 156104 & 32864 & 32864 &37.12$\pm$0.05 &45.72$\pm$2.42 &24.19$\pm$0.96 &28.00$\pm$0.89  \\
FractureMNIST3D & Hard & None & 156104 & 15656 & 40320 &37.25$\pm$0.08 &37.12$\pm$0.43 &21.36$\pm$0.19 &22.85$\pm$0.41  \\
NoduleMNIST3D & Easy & Pos. & 176016 & 25080 & 47120 &48.12$\pm$0.18 &48.09$\pm$0.07 &26.67$\pm$0.99 &26.48$\pm$0.29  \\
NoduleMNIST3D & Easy & Vol. & 176016 & 37056 & 37056 &39.99$\pm$0.31 &44.98$\pm$4.37 &23.88$\pm$0.16 &27.82$\pm$0.16  \\
NoduleMNIST3D & Easy & None & 176016 & 25080 & 52080 &40.19$\pm$0.05 &40.24$\pm$0.98 &22.99$\pm$0.49 &23.06$\pm$1.05  \\
NoduleMNIST3D & Hard & Pos. & 176016 & 25080 & 47120 &43.13$\pm$0.36 &43.57$\pm$0.37 &23.21$\pm$0.27 &23.60$\pm$0.31  \\
NoduleMNIST3D & Hard & Vol. & 176016 & 37056 & 37056 &36.64$\pm$0.06 &46.31$\pm$0.33 &23.68$\pm$0.58 &26.82$\pm$0.43  \\
NoduleMNIST3D & Hard & None & 176016 & 25080 & 52080 &36.08$\pm$0.34 &35.53$\pm$0.32 &18.87$\pm$0.23 &21.80$\pm$0.17  \\
VesselMNIST3D & Easy & Pos. & 202920 & 29032 & 58064 &52.66$\pm$0.49 &52.27$\pm$0.27 &33.22$\pm$0.28 &33.43$\pm$0.35 \\
VesselMNIST3D & Easy & Vol. & 202920 & 42720 & 42720 &41.45$\pm$0.52 &50.48$\pm$0.45 &32.23$\pm$1.01 &41.45$\pm$0.27  \\
VesselMNIST3D & Easy & None & 202920 & 29032 & 64176 &44.64$\pm$0.66 &45.47$\pm$0.84 &24.24$\pm$0.49 &24.17$\pm$0.35  \\
VesselMNIST3D & Hard & Pos. & 202920 & 29032 & 58064 &48.43$\pm$0.43 &49.43$\pm$0.16  &29.00$\pm$1.12 &29.31$\pm$0.35  \\
VesselMNIST3D & Hard & Vol. & 202920 & 42720 & 42720 &41.95$\pm$0.38 &48.92$\pm$0.65  &30.8$\pm$0.99 &35.05$\pm$0.42  \\
VesselMNIST3D & Hard & None & 202920 & 29032 & 64176 &41.08$\pm$0.28 &39.06$\pm$0.03 &19.72$\pm$0.35 &22.61$\pm$0.23  \\ \hline
\end{tabular}}
\caption{Comparison of k-5 accuracy (\%) with standard deviation among different dataset splitting methods.
Details of our three methods \textit{Pos.},\textit{Vol.} and \textit{None} can be found in Section~\ref{data_split}.
The slice-gathering method has easy and hard ones, details are shown in \ref{slice_gather}.
}
\label{exp_q3}
\end{table}

To analyze how different dataset-splitting methods affect the performance of RegisMCAN, we choose OrganMNIST3D, FractureMNIST3D, NoduleMNIST3D, and VesselMNIST3D as their train-test splitting is open-sourced. 
We create three different dataset-splitting methods. 
The first method \textit{Pos.} follows the original train-test splitting, and extracts the 2D tissue from the same position of each 3D organ volume data. This method will simulate the situation that physicians extract the 2D tissue from new patients. The patients have never been seen by RegisMCAN before, but the extracted position is exactly the same as the training samples.
The second method \textit{Vol.} combines all the 3D organ volume data into a whole set, then extracts 2D slices from different positions and splits them into training, validation and testing sets. This method simulates the situation that the patients' 3D organ have been recorded by RegisMCAN before and now physicians extract another 2D slice from his organ randomly again.
The third method \textit{None} simulates the most common case, meaning the model is trained with existing data, and used to justify a totally new patient with randomly extracted 2D tissue.

From Table~\ref{exp_q3} we find that method \textit{Vol.} always has the best performance and the method \textit{None} is the worst compared to the other variants. 
This is because although method \textit{Pos.} extracts the 2D tissue from the same position, due to the change of 3D volume data, the features of 2D tissue slice still change, resulting a worse result compared to \textit{Vol.}, whose 3D organ volume features are the same.
But for \textit{Pos.} method, its 2D slices are extracted from the same position, so they are still similar and easier for the RegisMCAN to learn, that's why its performance is better than \textit{None} method.

\subsection{Slice Gathering Effect}
\label{slice_gather}

To analyze how different slice-gathering methods affect the performance of RegisMCAN, we create two slice gathering methods. The \textit{Easy} one means that the slices are only extracted from three special views: coronal, sagittal and axial.
For the \textit{Hard} one, it is the normal case that the 2D tissues are randomly extracted from various perspectives.

In each Table~\ref{exp_q1}~\ref{exp_q2}~\ref{exp_q3}, we add the slice gathering comparison. The result shows that in most cases, the \textit{Easy} method will ease the task difficulty and helps RegisMCAN get better performance. However, there is a special case for ADNI-AD and ADNI-ALL in Table~\ref{exp_q2}. 
We infer that Alzheimer's Disease may lead to atrophy in the human brain, and thus many 2D tissue slices in \textit{Easy} samples may lose the features, resulting in the wrong result.
\section{Conclusions}

We figure out a new histology-to-organ registration problem which is crucial in next-generation biomedical research.
To solve this problem, we present {\model}, a benchmark dataset for multimodal tissue-organ registration and a pipeline for creating such type of benchmark. 
Then we present RegisMCAN, a possible VQA-based method to solve this problem by training with {\model}. 
RegisMCAN demonstrates convincing results in solving such a complex problem.
This will help physicians to better understand the basis of the radiologic signal and guide disease prediction based on non-invasive radiology studies.

While we believe that RegisMCAN represents a significant step towards solving the multimodal tissue-organ registration problem, we note that {\model} dataset and RegisMCAN still have some limitations. We discuss the limitations in Section~\ref{appendix}. Future work is directed toward improving quality and
reliability of {\model} dataset and RegisMCAN. We hope the {\model} can inspire more researchers to solve this real-world biomedical problem utilizing the power of deep learning.

\newpage
\appendix
\section{Appendix}
\label{appendix}
\subsection{Discussions of ATOM and RegisMCAN}

\textbf{Limitations of {\model}.} 
(1) Image resolution: With the development of hard devices, machine learning research can utilize GPUs with large memories such as 80G. However, 3D high-resolution volume data is still too large to be put into such GPUs. In {\model}, although we can create image with any resolution, we only build the low-resolution dataset, and scale down the 2D tissue slice at the same proportion now. This makes {\model} more friendly to machine learning sides but keeps a gap from the real-world scenario.
(2) Tissue-Slice problem: For machine learning purposes, {\model} collects 2D slices directly from the 3D organ data. However, in reality, the tissue is usually directly extracted from the human organ. 

\textbf{Limitations of RegisMCAN.} 
Precaution is required when utilizing the RegisMCAN model in practice: 
(1) Domain specificity: Although RegisMCAN can also be applied to other image registration problems, it was originally designed for the biomedical domain, and its performance may not be as effective in other domains. We welcome other researchers to test RegisMCAN on other applications, however, in this paper, we only show its capacity in the tissue-organ problem.
(2) Reliability: Like other AI models, the reliability of RegisMCAN is subject to the quality and quantity of the training data. There is always a possibility that it may not generalize well to certain types of images not covered in the training data. We strongly suggest users to double-check the results, and consider them as the preliminary responses that can be revised with expert knowledge and human judgment.
(3) Dependency on input quality: The quality of LLaVA-Med’s responses depends on the quality of the input data (2D tissue slice and 3D organ volume). Inaccurate or incomplete input data can lead to suboptimal assistance.
(4) Memory usage: While dealing with 3D images, the memory used by RegisMCAN will 
experience explosive growth of memory when increasing the 3D volume size. It will only use less than 2000MB while using {\model} as input. However, when we increase the 3D volume to size $\textbf{x}_{3D}$ $\in$ $\mathbb{R}^{40\times 40\times 40}$, it will cost more than 80G. So we also hope to propose new methods to decrease memory usage while keeping the performance in future.

\textbf{Future Works.} We encourage any researcher who is interested in this field to provide more kinds of volume data to {\model}. Also, future works can be done by developing imaging techniques to map the tissues from living organisms directly to images, this will make the data more in touch with facts.
With the development of memory efficiency methods, either a larger GPU or a novel sparsity method will help to deal with high-resolution data. We also encourage other researchers to try these methods for future works.

\end{document}